\documentclass{article}
\usepackage{spconf,amsmath,graphicx,hyperref}
\usepackage{amsmath}
\usepackage{amssymb}
\usepackage{subfig}
\usepackage{booktabs}       
\usepackage{multirow}       
\usepackage{makecell}       
\usepackage{graphicx}       
\usepackage{xcolor}         
\usepackage{threeparttable} 

\usepackage{etoolbox}

\apptocmd{\thebibliography}{
    \small
    \setlength{\itemsep}{-0.7pt}
    \setlength{\parskip}{0pt}
}{}{}

\title{Fewer yet critical: Reducing Redundant Token Dependencies for Transformer-based Time Series Forecasting}
\name{Jianqi Zhang$^{1,2}$, Yuchan Liu$^{1,2}$, Zeen Song$^{3}$, Yuefei Li$^{4}$ and Fanjiang Xu$^{1}$}

\address{
National Key Laboratory of Space Integrated Information Systems, Institute of Software, Chinese \\Academy of Sciences, Beijing, China $^1$;
University of Chinese Academy of Sciences, Beijing, China $^2$;  \\
Digital-Intelligent Operations Center, Guizhou Power Grid Co., Ltd., Guizhou, China $^3$;\\
College of Information Engineering, Hefei University of Economics$^4$.
}
\begin{document}
%
\maketitle
\begin{abstract}
Time series forecasting (TSF) is important in real-world applications. Recently, Transformer-based methods have achieved strong performance by modeling token dependencies through attention mechanisms. However, existing methods are usually trained mainly with prediction error losses, which may cause models to exploit both critical and redundant token dependencies. Such redundant dependencies can introduce irrelevant information and weaken generalization. To address this issue, we propose a simple yet effective token dependency selection strategy. Specifically, by jointly introducing the attention entropy constraint and prediction error constraint, the model can identify fewer but more critical inter-token dependencies and perform forecasting based on them, thus avoiding the interference of redundant dependencies. The proposed method can be easily extended to various Transformer-based TSF models. Experiments on multiple TSF datasets demonstrate its effectiveness.
\end{abstract}
\section{Introduction}
\label{sec:intro}
Time series forecasting (TSF) plays a crucial role in various real-world applications, such as weather prediction \cite{lee2025timecap,siahaan2023time}, energy planning \cite{boussif2024improving,novo2022planning}, and traffic flow forecasting \cite{fang2023stwave+,li2022dmgan}. Recently, Transformer-based methods have emerged as a dominant approach for TSF with strong predictive performance \cite{zhang2026enhancing, saravana2026transformers}. Typically, such methods first divide the input series into basic units called tokens. Then, the model uses the attention mechanism to model token dependencies, represented by attention scores, and aggregates information accordingly for prediction \cite{zhang2024not, zhang2026closing}.

\begin{figure}[htbp]
    \centering
    \includegraphics[width=1\linewidth]{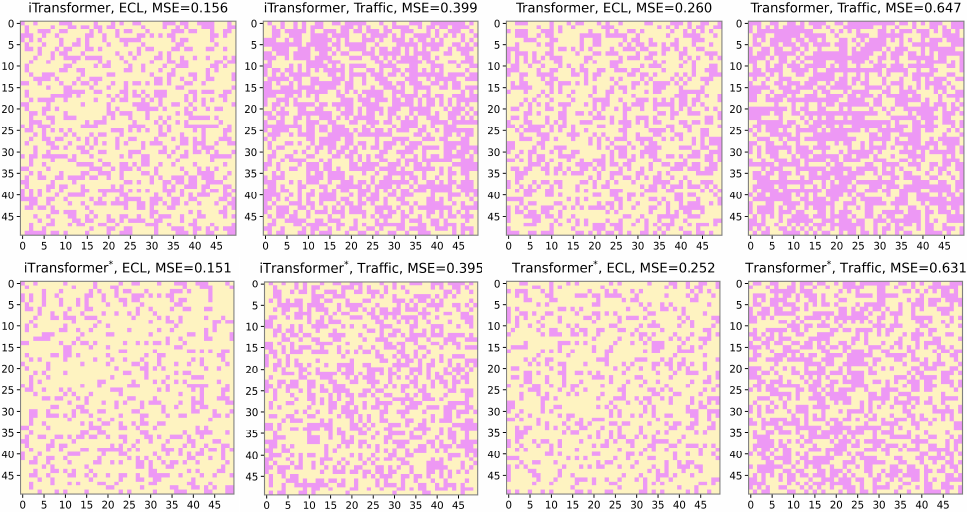}
    \caption{
    Visualization of the token dependency removal experiment. The method, dataset, and MSE are shown above each subfigure. $^{*}$ denotes the variant with the first-layer softmax temperature set to 0.8. The x- and y-axes denote token indices. Purple indicates dependencies whose removal improves performance, while beige indicates the opposite. For clarity, at most 50 tokens are shown.}
    \label{fig_motivate}
\end{figure}

However, existing Transformer-based TSF methods usually optimize model parameters mainly with prediction error losses, such as the Mean Squared Error (MSE) loss \cite{wang2026ri}. Under this training paradigm, as long as certain token dependencies help reduce the training loss, the model may assign them higher attention scores and exploit them for prediction. However, these dependencies are not necessarily critical to forecasting, and some of them may be redundant. Such redundant dependencies may arise from occasional abnormal fluctuations, measurement noise, or unstable short-term correlations in the training data \cite{zhou2024card, liu2022non}, making them unable to reflect stable and generalizable forecasting patterns. Without explicit constraints, the model can easily learn to rely on both critical and redundant dependencies during training. When facing unseen test data, this prediction pattern may introduce additional noise through redundant dependencies, thereby weakening forecasting performance.

To validate the above issue, we first conduct a dependency removal experiment on a trained Transformer-based TSF model. Specifically, we first train a standard TSF model. Then, for each pairwise token dependency in the first layer, we individually set its corresponding attention score to zero and evaluate the model on 500 randomly selected test samples. We record the change in MSE before and after removing each dependency, and visualize the results in the first row of Fig.\ref{fig_motivate}. Dependencies whose removal reduces the MSE are marked in purple, while the others are marked in beige. The experimental results show that many token dependencies can be removed without degrading model performance. This phenomenon indicates that existing models indeed exploit many redundant token dependencies during prediction, which can negatively affect forecasting results.

Furthermore, we investigate whether constraining the model’s use of token dependencies during training can alleviate the above issue. To this end, we set the softmax temperature coefficient of the first attention layer to 0.8, which achieves the best performance in the hyperparameter sensitivity analysis, and retrain the model. A lower temperature coefficient makes the attention score distribution more concentrated and suppresses more low-weight dependencies to values close to zero, thereby encouraging the model to use fewer token dependencies for prediction. After training, we repeat the above dependency removal experiment. The results (the second row of Fig.\ref{fig_motivate}) show that, compared with the original model, the temperature-constrained model contains fewer redundant dependencies and achieves improved forecasting performance. This suggests that constraining the model to make accurate predictions with fewer dependencies during training can reduce its reliance on redundant dependencies to some extent, thereby improving forecasting performance.

The above discussion shows that Transformer-based TSF models should be explicitly constrained during training to use fewer but more critical token dependencies, thereby reducing the interference of redundant dependencies and improving generalization performance. However, the fixed temperature coefficient constraint lacks flexibility and cannot adapt well to different layers, samples, and data patterns, resulting in limited performance gains, as shown in Fig.\ref{fig_motivate}. Therefore, a more flexible adaptive constraint method is needed.

Based on this insight, we propose a training constraint method that encourages Transformer-based TSF models to focus on fewer yet more critical token dependencies for better forecasting. Specifically, we jointly introduce the attention entropy constraint and prediction error constraint during training. Attention entropy measures the dispersion of attention scores, with lower entropy indicating more concentrated attention. By minimizing attention entropy, the model learns more concentrated attention distributions and reduces dispersed focus on token relations. Meanwhile, by minimizing the prediction loss, we ensure that the retained high-weight dependencies can still support accurate forecasting. This joint optimization helps the model rely on fewer but more critical token dependencies for forecasting, thereby reducing the interference of redundant dependencies and ultimately improving performance. Extensive experiments on multiple TSF tasks demonstrate the effectiveness of our method. 

Our main contributions are as follows:
1) We empirically reveal that existing Transformer-based TSF models exploit many redundant token dependencies during forecasting.
2) We propose a simple yet effective token dependency selection strategy, which combines attention entropy constraints with prediction error constraints to encourage models to rely on fewer but more critical token dependencies.
3) The proposed method is model-agnostic and can be easily extended to various Transformer-based TSF models. Extensive experiments demonstrate its effectiveness and superiority.

\section{Related Works}
To reduce the computational cost of self-attention in long-sequence modeling, some TSF methods, such as LogSparse Transformer \cite{li2019enhancing}, Informer \cite{zhou2021informer}, and Infomaxformer \cite{tang2023infomaxformer}, employ fixed sparse attention mechanisms to improve modeling efficiency. In addition, ARCausal \cite{zhou2026causal} further considers redundant inter-variable correlations. However, it requires an additional causal discovery module for variables and focuses only on redundant dependencies among variable tokens, where each variable corresponds to one token. In contrast, our method does not aim to reduce the computational cost of self-attention, is not restricted to a specific token type, and requires neither predefined sparse attention patterns nor additional causal discovery modules. Therefore, our work is fundamentally different from these existing sparse attention methods in both research objective and scope of applicability.

\begin{figure}[htbp]
    \centering
    \includegraphics[width=1\linewidth]{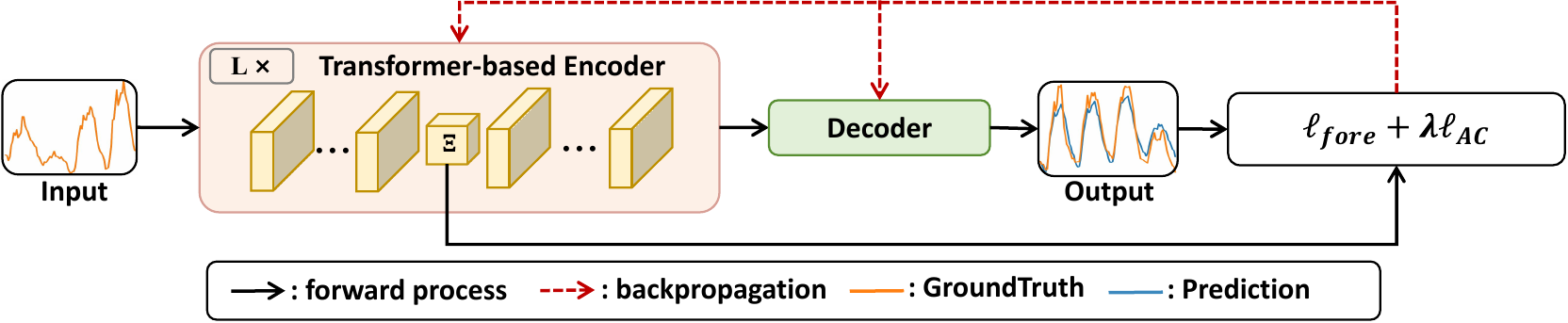}
    \caption{The overall framework of the proposed method. }
    \label{fig_model}
\end{figure} 

\section{Methodology}
\label{sec:method}
In this section, we introduce our method. The core idea of our method is to constrain attention distributions during training, encouraging accurate predictions with fewer token dependencies. The overall procedure of our method is shown in Fig.\ref{fig_model}.

\subsection{Problem Formulation}
\textbf{TSF task: }Given a historical multivariate time series $\mathbf{X}\in \mathbb{R}^{L_{in}\times C}$, where $L_{in}$ denotes the input length and $C$ denotes the number of variables, the goal of time series forecasting is to predict the future series $\mathbf{Y}\in\mathbb{R}^{L_{out}\times C}$ of length $L_{out}$.

\textbf{Transformer-based TSF methods: }
These methods leverage the attention mechanism to extract token dependencies for prediction. Specifically, given the input $X$, these methods first divide it into subsequences. Then an embedding function $f_{emb}$ is used to convert each subsequence into a basic unit called ``token''. After obtaining all the input tokens $Tok=\{Tok_1, Tok_2, ...\}$, the model uses three linear layers to transform $Tok$ into Query ($Q$), Key ($K$), and Value ($V$). Query ($Q$) and Key ($K$) are used to compute the attention map ($\Xi$). The attention map is a two-dimensional matrix whose elements are called attention scores, which represent the \textbf{token dependencies}. The calculation method for the attention map ($\Xi$) is as follows:
\begin{equation}
\label{eq_1}
\Xi = \mathcal{S}(QK^{\top}/\sqrt{D}), 
\end{equation}
where $D$ is the dimension of tokens, $\mathcal{S}(\cdot)$ is the row-wise softmax function. Then, the model multiplies $\Xi$ with $V$ and feeds the result into an FFN to obtain new tokens:
\begin{equation}
\label{eq_2}
Tok^{new} = FFN(\Xi V), 
\end{equation}
This process (Eq.\ref{eq_1}, \ref{eq_2}) is repeated multiple times to generate the final token representation $Tok^{f}=\{Tok^{f}_1, ..., Tok^{f}_N\}$. The final prediction is obtained by decoding $Tok^{f}$.

\subsection{Attention Entropy Constraint}
To reduce the model's reliance on redundant token dependencies for forecasting, we introduce an attention entropy constraint during training. This constraint encourages the attention distribution to become more concentrated, thereby enabling the model to make predictions based on fewer token dependencies. Specifically, each row of the attention map ($\Xi$) in Eq.\ref{eq_1} constitutes a complete probability distribution, where the magnitude of each probability value reflects the degree to which the model relies on the corresponding token dependency. We define the attention entropy of $\Xi$ at the $l$-th layer as the mean entropy of its row-wise distributions:
\begin{equation}
\label{eq_entropy}
\mathcal{H}^{l}
=
-\frac{1}{N\log N}
\sum_{i=1}^{N}
\sum_{j=1}^{N}
\Xi^{l}_{i,j}
\log\left(\Xi^{l}_{i,j}+\epsilon\right),
\end{equation}
where $N$ is the size of $\Xi$, $\epsilon$=1e-12 is a small positive constant used to avoid $\log 0$. We use the mean of $\mathcal{H}^{l}$ across all encoder layers as the overall attention entropy constraint: $\ell_{AC}=\frac{1}{L}\sum_{l=1}^{L}\mathcal{H}^{l}$, where $L$ is the number of encoder layers in the model. Attention entropy reflects the dispersion of attention weights: higher entropy indicates reliance on more token dependencies, while lower entropy indicates reliance on fewer ones. Thus, minimizing $\ell_{AC}$ encourages the model to concentrate its attention on fewer token dependencies.

\subsection{Joint Optimization}
The overall optimization objective in this paper is as follows:
\begin{equation}
\label{eq_object}
\ell_{total}=\ell_{fore}+ \lambda \ell_{AC}
\end{equation}
where $\ell_{fore}$ denotes the MSE loss and $\lambda$ is a hyperparameter. The two terms in $\ell_{total}$ play complementary roles. The former encourages the model to make predictions as accurately as possible, while the latter encourages it to rely on as few token dependencies as possible. Through joint optimization, the two objectives encourage the model to make predictions based on fewer but more crucial token dependencies, thereby reducing the interference of redundant dependencies and improving forecasting performance.

\begin{figure}[htbp]
    \centering
    \includegraphics[width=1\linewidth]{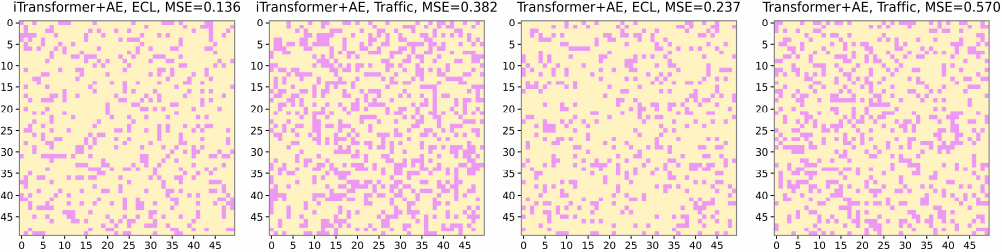}
    \caption{Visualization of the token dependency removal experiment. +AE means adding our Attention Entropy Constraint. }
    \label{visualize}
\end{figure} 

\begin{table*}[t]
\caption{Full results for the TSF task. The look-back length for all baseline models is 96. The results of our method are averaged over five random seeds. The best results are highlighted in red, and the second-best results are highlighted in blue.}
\vskip -0.0in
\renewcommand{\arraystretch}{0.86}
\centering
\resizebox{1.55\columnwidth}{!}{%
\begin{threeparttable}
\begin{small}
\setlength{\tabcolsep}{1pt}
\begin{tabular}{c|c|cc|cc|cc|cc|cc|cc|cc|cc|cc|cc|cc|cc|cc|cc}
\toprule
\multicolumn{2}{c}{\multirow{2}{*}{Models}} & \multicolumn{2}{c}{\rotatebox{0}{\scalebox{0.80}{\textbf{iTransformer}}}} & \multicolumn{2}{c}{\rotatebox{0}{\scalebox{0.80}{iTransformer}}} & \multicolumn{2}{c}{\rotatebox{0}{\scalebox{0.80}{\textbf{PatchTST}}}} & \multicolumn{2}{c}{\rotatebox{0}{\scalebox{0.80}{PatchTST}}} & \multicolumn{2}{c}{\rotatebox{0}{\scalebox{0.80}{\textbf{Transformer}}}} & \multicolumn{2}{c}{\rotatebox{0}{\scalebox{0.80}{Transformer}}} & \multicolumn{2}{c}{\rotatebox{0}{\scalebox{0.80}{CycleNet}}} & \multicolumn{2}{c}{\rotatebox{0}{\scalebox{0.80}{S-Mamba}}} & \multicolumn{2}{c}{\rotatebox{0}{\scalebox{0.80}{Crossformer}}} & \multicolumn{2}{c}{\rotatebox{0}{\scalebox{0.80}{DSformer}}} & \multicolumn{2}{c}{\rotatebox{0}{\scalebox{0.80}{TiDE}}} & \multicolumn{2}{c}{\rotatebox{0}{\scalebox{0.80}{DPAD}}} & \multicolumn{2}{c}{\rotatebox{0}{\scalebox{0.80}{ARCausal}}} & \multicolumn{2}{c}{\rotatebox{0}{\scalebox{0.80}{Informer}}} \\
\multicolumn{2}{c}{} & \multicolumn{2}{c}{\scalebox{0.76}{\textbf{(+AE)}}} & \multicolumn{2}{c}{\scalebox{0.76}{(2024)}} & \multicolumn{2}{c}{\scalebox{0.76}{\textbf{(+AE)}}} & \multicolumn{2}{c}{\scalebox{0.76}{(2023)}} & \multicolumn{2}{c}{\scalebox{0.76}{\textbf{(+AE)}}} & \multicolumn{2}{c}{\scalebox{0.76}{(2017)}} & \multicolumn{2}{c}{\scalebox{0.76}{/Linear (2024)}} & \multicolumn{2}{c}{\scalebox{0.76}{(2025)}} & \multicolumn{2}{c}{\scalebox{0.76}{(2023)}} & \multicolumn{2}{c}{\scalebox{0.76}{(2023)}} & \multicolumn{2}{c}{\scalebox{0.76}{(2023)}} & \multicolumn{2}{c}{\scalebox{0.76}{(2026)}} & \multicolumn{2}{c}{\scalebox{0.76}{(2026)}} & \multicolumn{2}{c}{\scalebox{0.76}{(2021)}} \\
\cmidrule(lr){3-4} \cmidrule(lr){5-6} \cmidrule(lr){7-8} \cmidrule(lr){9-10} \cmidrule(lr){11-12} \cmidrule(lr){13-14} \cmidrule(lr){15-16} \cmidrule(lr){17-18} \cmidrule(lr){19-20} \cmidrule(lr){21-22} \cmidrule(lr){23-24} \cmidrule(lr){25-26} \cmidrule(lr){27-28} \cmidrule(lr){29-30}
\multicolumn{2}{c}{Metric} & \scalebox{0.78}{MSE} & \scalebox{0.78}{MAE} & \scalebox{0.78}{MSE} & \scalebox{0.78}{MAE} & \scalebox{0.78}{MSE} & \scalebox{0.78}{MAE} & \scalebox{0.78}{MSE} & \scalebox{0.78}{MAE} & \scalebox{0.78}{MSE} & \scalebox{0.78}{MAE} & \scalebox{0.78}{MSE} & \scalebox{0.78}{MAE} & \scalebox{0.78}{MSE} & \scalebox{0.78}{MAE} & \scalebox{0.78}{MSE} & \scalebox{0.78}{MAE} & \scalebox{0.78}{MSE} & \scalebox{0.78}{MAE} & \scalebox{0.78}{MSE} & \scalebox{0.78}{MAE} & \scalebox{0.78}{MSE} & \scalebox{0.78}{MAE} & \scalebox{0.78}{MSE} & \scalebox{0.78}{MAE} & \scalebox{0.78}{MSE} & \scalebox{0.78}{MAE} & \scalebox{0.78}{MSE} & \scalebox{0.78}{MAE} \\
\toprule
\multirow{5}{*}{\rotatebox{90}{\scalebox{0.95}{ETTh2}}} & \scalebox{0.78}{96} & \scalebox{0.78}{\textcolor{blue}{0.293}} & \scalebox{0.78}{\textcolor{blue}{0.341}} & \scalebox{0.78}{0.297} & \scalebox{0.78}{0.347} & \scalebox{0.78}{0.301} & \scalebox{0.78}{0.346} & \scalebox{0.78}{0.302} & \scalebox{0.78}{0.349} & \scalebox{0.78}{0.455} & \scalebox{0.78}{0.442} & \scalebox{0.78}{0.470} & \scalebox{0.78}{0.461} & \scalebox{0.78}{\textcolor{red}{0.285}} & \scalebox{0.78}{\textcolor{red}{0.335}} & \scalebox{0.78}{0.296} & \scalebox{0.78}{0.348} & \scalebox{0.78}{0.745} & \scalebox{0.78}{0.584} & \scalebox{0.78}{0.319} & \scalebox{0.78}{0.359} & \scalebox{0.78}{0.400} & \scalebox{0.78}{0.440} & \scalebox{0.78}{\textcolor{blue}{0.293}} & \scalebox{0.78}{0.345} & \scalebox{0.78}{0.295} & \scalebox{0.78}{0.345} & \scalebox{0.78}{3.755} & \scalebox{0.78}{1.525} \\
 & \scalebox{0.78}{192} & \scalebox{0.78}{0.375} & \scalebox{0.78}{\textcolor{red}{0.390}} & \scalebox{0.78}{0.380} & \scalebox{0.78}{0.400} & \scalebox{0.78}{0.381} & \scalebox{0.78}{0.397} & \scalebox{0.78}{0.388} & \scalebox{0.78}{0.400} & \scalebox{0.78}{0.569} & \scalebox{0.78}{0.502} & \scalebox{0.78}{0.601} & \scalebox{0.78}{0.532} & \scalebox{0.78}{\textcolor{red}{0.373}} & \scalebox{0.78}{\textcolor{blue}{0.391}} & \scalebox{0.78}{0.376} & \scalebox{0.78}{0.396} & \scalebox{0.78}{0.877} & \scalebox{0.78}{0.656} & \scalebox{0.78}{0.400} & \scalebox{0.78}{0.410} & \scalebox{0.78}{0.528} & \scalebox{0.78}{0.509} & \scalebox{0.78}{\textcolor{blue}{0.374}} & \scalebox{0.78}{0.393} & \scalebox{0.78}{0.375} & \scalebox{0.78}{0.394} & \scalebox{0.78}{5.602} & \scalebox{0.78}{1.931} \\
 & \scalebox{0.78}{336} & \scalebox{0.78}{0.423} & \scalebox{0.78}{0.431} & \scalebox{0.78}{0.428} & \scalebox{0.78}{0.432} & \scalebox{0.78}{0.421} & \scalebox{0.78}{0.434} & \scalebox{0.78}{0.426} & \scalebox{0.78}{0.433} & \scalebox{0.78}{0.641} & \scalebox{0.78}{0.553} & \scalebox{0.78}{0.676} & \scalebox{0.78}{0.575} & \scalebox{0.78}{0.421} & \scalebox{0.78}{0.433} & \scalebox{0.78}{0.424} & \scalebox{0.78}{0.431} & \scalebox{0.78}{1.043} & \scalebox{0.78}{0.731} & \scalebox{0.78}{0.462} & \scalebox{0.78}{0.451} & \scalebox{0.78}{0.643} & \scalebox{0.78}{0.571} & \scalebox{0.78}{\textcolor{red}{0.416}} & \scalebox{0.78}{\textcolor{red}{0.425}} & \scalebox{0.78}{\textcolor{blue}{0.419}} & \scalebox{0.78}{\textcolor{blue}{0.429}} & \scalebox{0.78}{4.721} & \scalebox{0.78}{1.835} \\
 & \scalebox{0.78}{720} & \scalebox{0.78}{\textcolor{red}{0.421}} & \scalebox{0.78}{\textcolor{blue}{0.440}} & \scalebox{0.78}{0.427} & \scalebox{0.78}{0.445} & \scalebox{0.78}{0.425} & \scalebox{0.78}{0.444} & \scalebox{0.78}{0.431} & \scalebox{0.78}{0.446} & \scalebox{0.78}{0.646} & \scalebox{0.78}{0.561} & \scalebox{0.78}{0.676} & \scalebox{0.78}{0.592} & \scalebox{0.78}{0.453} & \scalebox{0.78}{0.458} & \scalebox{0.78}{0.426} & \scalebox{0.78}{0.444} & \scalebox{0.78}{1.104} & \scalebox{0.78}{0.763} & \scalebox{0.78}{0.457} & \scalebox{0.78}{0.463} & \scalebox{0.78}{0.874} & \scalebox{0.78}{0.679} & \scalebox{0.78}{\textcolor{blue}{0.422}} & \scalebox{0.78}{\textcolor{red}{0.435}} & \scalebox{0.78}{0.426} & \scalebox{0.78}{0.443} & \scalebox{0.78}{3.647} & \scalebox{0.78}{1.625} \\
\cmidrule(lr){2-30}
 & \scalebox{0.78}{Avg} & \scalebox{0.78}{\textcolor{blue}{0.378}} & \scalebox{0.78}{\textcolor{blue}{0.401}} & \scalebox{0.78}{0.383} & \scalebox{0.78}{0.406} & \scalebox{0.78}{0.382} & \scalebox{0.78}{0.405} & \scalebox{0.78}{0.387} & \scalebox{0.78}{0.407} & \scalebox{0.78}{0.578} & \scalebox{0.78}{0.514} & \scalebox{0.78}{0.606} & \scalebox{0.78}{0.540} & \scalebox{0.78}{0.383} & \scalebox{0.78}{0.404} & \scalebox{0.78}{0.380} & \scalebox{0.78}{0.405} & \scalebox{0.78}{0.942} & \scalebox{0.78}{0.683} & \scalebox{0.78}{0.410} & \scalebox{0.78}{0.421} & \scalebox{0.78}{0.611} & \scalebox{0.78}{0.550} & \scalebox{0.78}{\textcolor{red}{0.376}} & \scalebox{0.78}{\textcolor{red}{0.400}} & \scalebox{0.78}{0.379} & \scalebox{0.78}{0.403} & \scalebox{0.78}{4.431} & \scalebox{0.78}{1.729} \\
\midrule
\multirow{5}{*}{\rotatebox{90}{\scalebox{0.95}{ECL}}} & \scalebox{0.78}{96} & \scalebox{0.78}{\textcolor{red}{0.135}} & \scalebox{0.78}{\textcolor{red}{0.226}} & \scalebox{0.78}{0.148} & \scalebox{0.78}{0.240} & \scalebox{0.78}{0.190} & \scalebox{0.78}{0.281} & \scalebox{0.78}{0.195} & \scalebox{0.78}{0.285} & \scalebox{0.78}{0.232} & \scalebox{0.78}{0.320} & \scalebox{0.78}{0.260} & \scalebox{0.78}{0.358} & \scalebox{0.78}{0.141} & \scalebox{0.78}{\textcolor{blue}{0.234}} & \scalebox{0.78}{\textcolor{blue}{0.139}} & \scalebox{0.78}{0.235} & \scalebox{0.78}{0.219} & \scalebox{0.78}{0.314} & \scalebox{0.78}{0.173} & \scalebox{0.78}{0.269} & \scalebox{0.78}{0.237} & \scalebox{0.78}{0.329} & \scalebox{0.78}{0.147} & \scalebox{0.78}{\textcolor{blue}{0.234}} & \scalebox{0.78}{0.145} & \scalebox{0.78}{\textcolor{blue}{0.234}} & \scalebox{0.78}{0.274} & \scalebox{0.78}{0.368} \\
 & \scalebox{0.78}{192} & \scalebox{0.78}{\textcolor{red}{0.152}} & \scalebox{0.78}{\textcolor{blue}{0.248}} & \scalebox{0.78}{0.162} & \scalebox{0.78}{0.253} & \scalebox{0.78}{0.193} & \scalebox{0.78}{0.281} & \scalebox{0.78}{0.199} & \scalebox{0.78}{0.289} & \scalebox{0.78}{0.241} & \scalebox{0.78}{0.332} & \scalebox{0.78}{0.266} & \scalebox{0.78}{0.367} & \scalebox{0.78}{\textcolor{blue}{0.155}} & \scalebox{0.78}{\textcolor{red}{0.247}} & \scalebox{0.78}{0.159} & \scalebox{0.78}{0.255} & \scalebox{0.78}{0.231} & \scalebox{0.78}{0.322} & \scalebox{0.78}{0.183} & \scalebox{0.78}{0.280} & \scalebox{0.78}{0.236} & \scalebox{0.78}{0.330} & \scalebox{0.78}{0.162} & \scalebox{0.78}{0.251} & \scalebox{0.78}{0.160} & \scalebox{0.78}{\textcolor{red}{0.247}} & \scalebox{0.78}{0.296} & \scalebox{0.78}{0.386} \\
 & \scalebox{0.78}{336} & \scalebox{0.78}{\textcolor{red}{0.166}} & \scalebox{0.78}{\textcolor{red}{0.262}} & \scalebox{0.78}{0.178} & \scalebox{0.78}{0.269} & \scalebox{0.78}{0.208} & \scalebox{0.78}{0.300} & \scalebox{0.78}{0.215} & \scalebox{0.78}{0.305} & \scalebox{0.78}{0.250} & \scalebox{0.78}{0.336} & \scalebox{0.78}{0.280} & \scalebox{0.78}{0.375} & \scalebox{0.78}{\textcolor{blue}{0.172}} & \scalebox{0.78}{0.264} & \scalebox{0.78}{0.176} & \scalebox{0.78}{0.272} & \scalebox{0.78}{0.246} & \scalebox{0.78}{0.337} & \scalebox{0.78}{0.203} & \scalebox{0.78}{0.297} & \scalebox{0.78}{0.249} & \scalebox{0.78}{0.344} & \scalebox{0.78}{\textcolor{blue}{0.172}} & \scalebox{0.78}{0.271} & \scalebox{0.78}{0.173} & \scalebox{0.78}{\textcolor{blue}{0.263}} & \scalebox{0.78}{0.300} & \scalebox{0.78}{0.394} \\
 & \scalebox{0.78}{720} & \scalebox{0.78}{\textcolor{red}{0.199}} & \scalebox{0.78}{0.298} & \scalebox{0.78}{0.225} & \scalebox{0.78}{0.317} & \scalebox{0.78}{0.253} & \scalebox{0.78}{0.334} & \scalebox{0.78}{0.256} & \scalebox{0.78}{0.337} & \scalebox{0.78}{0.270} & \scalebox{0.78}{0.346} & \scalebox{0.78}{0.302} & \scalebox{0.78}{0.386} & \scalebox{0.78}{0.210} & \scalebox{0.78}{\textcolor{blue}{0.296}} & \scalebox{0.78}{0.204} & \scalebox{0.78}{0.298} & \scalebox{0.78}{0.280} & \scalebox{0.78}{0.363} & \scalebox{0.78}{0.259} & \scalebox{0.78}{0.340} & \scalebox{0.78}{0.284} & \scalebox{0.78}{0.373} & \scalebox{0.78}{\textcolor{blue}{0.201}} & \scalebox{0.78}{\textcolor{red}{0.289}} & \scalebox{0.78}{0.210} & \scalebox{0.78}{0.297} & \scalebox{0.78}{0.373} & \scalebox{0.78}{0.439} \\
\cmidrule(lr){2-30}
 & \scalebox{0.78}{Avg} & \scalebox{0.78}{\textcolor{red}{0.163}} & \scalebox{0.78}{\textcolor{red}{0.259}} & \scalebox{0.78}{0.178} & \scalebox{0.78}{0.270} & \scalebox{0.78}{0.211} & \scalebox{0.78}{0.299} & \scalebox{0.78}{0.216} & \scalebox{0.78}{0.304} & \scalebox{0.78}{0.248} & \scalebox{0.78}{0.334} & \scalebox{0.78}{0.277} & \scalebox{0.78}{0.372} & \scalebox{0.78}{\textcolor{blue}{0.169}} & \scalebox{0.78}{\textcolor{blue}{0.260}} & \scalebox{0.78}{0.170} & \scalebox{0.78}{0.265} & \scalebox{0.78}{0.244} & \scalebox{0.78}{0.334} & \scalebox{0.78}{0.204} & \scalebox{0.78}{0.297} & \scalebox{0.78}{0.252} & \scalebox{0.78}{0.344} & \scalebox{0.78}{0.170} & \scalebox{0.78}{0.261} & \scalebox{0.78}{0.172} & \scalebox{0.78}{\textcolor{blue}{0.260}} & \scalebox{0.78}{0.311} & \scalebox{0.78}{0.397} \\
\midrule
\multirow{5}{*}{\rotatebox{90}{\scalebox{0.95}{Traffic}}} & \scalebox{0.78}{96} & \scalebox{0.78}{\textcolor{red}{0.382}} & \scalebox{0.78}{\textcolor{blue}{0.260}} & \scalebox{0.78}{0.395} & \scalebox{0.78}{0.268} & \scalebox{0.78}{0.532} & \scalebox{0.78}{0.353} & \scalebox{0.78}{0.544} & \scalebox{0.78}{0.359} & \scalebox{0.78}{0.579} & \scalebox{0.78}{0.319} & \scalebox{0.78}{0.647} & \scalebox{0.78}{0.357} & \scalebox{0.78}{0.480} & \scalebox{0.78}{0.314} & \scalebox{0.78}{\textcolor{red}{0.382}} & \scalebox{0.78}{0.261} & \scalebox{0.78}{0.522} & \scalebox{0.78}{0.290} & \scalebox{0.78}{0.529} & \scalebox{0.78}{0.370} & \scalebox{0.78}{0.805} & \scalebox{0.78}{0.493} & \scalebox{0.78}{0.392} & \scalebox{0.78}{0.268} & \scalebox{0.78}{\textcolor{blue}{0.389}} & \scalebox{0.78}{\textcolor{red}{0.256}} & \scalebox{0.78}{0.719} & \scalebox{0.78}{0.391} \\
 & \scalebox{0.78}{192} & \scalebox{0.78}{\textcolor{blue}{0.403}} & \scalebox{0.78}{0.268} & \scalebox{0.78}{0.417} & \scalebox{0.78}{0.276} & \scalebox{0.78}{0.531} & \scalebox{0.78}{0.350} & \scalebox{0.78}{0.540} & \scalebox{0.78}{0.354} & \scalebox{0.78}{0.581} & \scalebox{0.78}{0.318} & \scalebox{0.78}{0.649} & \scalebox{0.78}{0.356} & \scalebox{0.78}{0.482} & \scalebox{0.78}{0.313} & \scalebox{0.78}{\textcolor{red}{0.396}} & \scalebox{0.78}{\textcolor{blue}{0.267}} & \scalebox{0.78}{0.530} & \scalebox{0.78}{0.293} & \scalebox{0.78}{0.533} & \scalebox{0.78}{0.366} & \scalebox{0.78}{0.756} & \scalebox{0.78}{0.474} & \scalebox{0.78}{0.407} & \scalebox{0.78}{0.271} & \scalebox{0.78}{0.408} & \scalebox{0.78}{\textcolor{red}{0.264}} & \scalebox{0.78}{0.696} & \scalebox{0.78}{0.379} \\
 & \scalebox{0.78}{336} & \scalebox{0.78}{\textcolor{red}{0.412}} & \scalebox{0.78}{\textcolor{blue}{0.274}} & \scalebox{0.78}{0.433} & \scalebox{0.78}{0.283} & \scalebox{0.78}{0.544} & \scalebox{0.78}{0.355} & \scalebox{0.78}{0.551} & \scalebox{0.78}{0.358} & \scalebox{0.78}{0.597} & \scalebox{0.78}{0.326} & \scalebox{0.78}{0.667} & \scalebox{0.78}{0.364} & \scalebox{0.78}{0.476} & \scalebox{0.78}{0.303} & \scalebox{0.78}{\textcolor{blue}{0.417}} & \scalebox{0.78}{0.276} & \scalebox{0.78}{0.558} & \scalebox{0.78}{0.305} & \scalebox{0.78}{0.545} & \scalebox{0.78}{0.370} & \scalebox{0.78}{0.762} & \scalebox{0.78}{0.477} & \scalebox{0.78}{\textcolor{blue}{0.417}} & \scalebox{0.78}{0.279} & \scalebox{0.78}{0.424} & \scalebox{0.78}{\textcolor{red}{0.272}} & \scalebox{0.78}{0.777} & \scalebox{0.78}{0.420} \\
 & \scalebox{0.78}{720} & \scalebox{0.78}{\textcolor{red}{0.441}} & \scalebox{0.78}{\textcolor{red}{0.289}} & \scalebox{0.78}{0.467} & \scalebox{0.78}{0.302} & \scalebox{0.78}{0.580} & \scalebox{0.78}{0.369} & \scalebox{0.78}{0.586} & \scalebox{0.78}{0.375} & \scalebox{0.78}{0.630} & \scalebox{0.78}{0.340} & \scalebox{0.78}{0.697} & \scalebox{0.78}{0.376} & \scalebox{0.78}{0.503} & \scalebox{0.78}{0.320} & \scalebox{0.78}{0.460} & \scalebox{0.78}{0.300} & \scalebox{0.78}{0.589} & \scalebox{0.78}{0.328} & \scalebox{0.78}{0.583} & \scalebox{0.78}{0.386} & \scalebox{0.78}{0.719} & \scalebox{0.78}{0.449} & \scalebox{0.78}{\textcolor{blue}{0.447}} & \scalebox{0.78}{0.293} & \scalebox{0.78}{0.459} & \scalebox{0.78}{\textcolor{blue}{0.290}} & \scalebox{0.78}{0.864} & \scalebox{0.78}{0.472} \\
\cmidrule(lr){2-30}
 & \scalebox{0.78}{Avg} & \scalebox{0.78}{\textcolor{red}{0.410}} & \scalebox{0.78}{\textcolor{blue}{0.273}} & \scalebox{0.78}{0.428} & \scalebox{0.78}{0.282} & \scalebox{0.78}{0.547} & \scalebox{0.78}{0.357} & \scalebox{0.78}{0.555} & \scalebox{0.78}{0.361} & \scalebox{0.78}{0.597} & \scalebox{0.78}{0.326} & \scalebox{0.78}{0.665} & \scalebox{0.78}{0.363} & \scalebox{0.78}{0.485} & \scalebox{0.78}{0.312} & \scalebox{0.78}{\textcolor{blue}{0.414}} & \scalebox{0.78}{0.276} & \scalebox{0.78}{0.550} & \scalebox{0.78}{0.304} & \scalebox{0.78}{0.548} & \scalebox{0.78}{0.373} & \scalebox{0.78}{0.760} & \scalebox{0.78}{0.473} & \scalebox{0.78}{0.416} & \scalebox{0.78}{0.278} & \scalebox{0.78}{0.420} & \scalebox{0.78}{\textcolor{red}{0.271}} & \scalebox{0.78}{0.764} & \scalebox{0.78}{0.415} \\
\midrule
\multirow{5}{*}{\rotatebox{90}{\scalebox{0.95}{Weather}}} & \scalebox{0.78}{96} & \scalebox{0.78}{\textcolor{red}{0.156}} & \scalebox{0.78}{\textcolor{red}{0.203}} & \scalebox{0.78}{0.174} & \scalebox{0.78}{0.214} & \scalebox{0.78}{0.174} & \scalebox{0.78}{0.215} & \scalebox{0.78}{0.177} & \scalebox{0.78}{0.218} & \scalebox{0.78}{0.359} & \scalebox{0.78}{0.387} & \scalebox{0.78}{0.395} & \scalebox{0.78}{0.427} & \scalebox{0.78}{0.170} & \scalebox{0.78}{0.216} & \scalebox{0.78}{0.165} & \scalebox{0.78}{0.210} & \scalebox{0.78}{\textcolor{blue}{0.158}} & \scalebox{0.78}{0.230} & \scalebox{0.78}{0.162} & \scalebox{0.78}{\textcolor{blue}{0.207}} & \scalebox{0.78}{0.202} & \scalebox{0.78}{0.261} & \scalebox{0.78}{0.170} & \scalebox{0.78}{0.208} & \scalebox{0.78}{0.169} & \scalebox{0.78}{0.208} & \scalebox{0.78}{0.300} & \scalebox{0.78}{0.384} \\
 & \scalebox{0.78}{192} & \scalebox{0.78}{\textcolor{blue}{0.209}} & \scalebox{0.78}{\textcolor{red}{0.249}} & \scalebox{0.78}{0.221} & \scalebox{0.78}{0.254} & \scalebox{0.78}{0.220} & \scalebox{0.78}{0.254} & \scalebox{0.78}{0.225} & \scalebox{0.78}{0.259} & \scalebox{0.78}{0.554} & \scalebox{0.78}{0.502} & \scalebox{0.78}{0.619} & \scalebox{0.78}{0.560} & \scalebox{0.78}{0.222} & \scalebox{0.78}{0.259} & \scalebox{0.78}{0.214} & \scalebox{0.78}{0.252} & \scalebox{0.78}{\textcolor{red}{0.206}} & \scalebox{0.78}{0.277} & \scalebox{0.78}{0.211} & \scalebox{0.78}{0.252} & \scalebox{0.78}{0.242} & \scalebox{0.78}{0.298} & \scalebox{0.78}{0.223} & \scalebox{0.78}{0.256} & \scalebox{0.78}{0.211} & \scalebox{0.78}{\textcolor{blue}{0.250}} & \scalebox{0.78}{0.598} & \scalebox{0.78}{0.544} \\
 & \scalebox{0.78}{336} & \scalebox{0.78}{\textcolor{red}{0.265}} & \scalebox{0.78}{\textcolor{red}{0.290}} & \scalebox{0.78}{0.278} & \scalebox{0.78}{0.296} & \scalebox{0.78}{0.275} & \scalebox{0.78}{0.294} & \scalebox{0.78}{0.278} & \scalebox{0.78}{0.297} & \scalebox{0.78}{0.617} & \scalebox{0.78}{0.533} & \scalebox{0.78}{0.689} & \scalebox{0.78}{0.594} & \scalebox{0.78}{0.275} & \scalebox{0.78}{0.296} & \scalebox{0.78}{0.274} & \scalebox{0.78}{0.297} & \scalebox{0.78}{0.272} & \scalebox{0.78}{0.335} & \scalebox{0.78}{\textcolor{blue}{0.267}} & \scalebox{0.78}{0.294} & \scalebox{0.78}{0.287} & \scalebox{0.78}{0.335} & \scalebox{0.78}{0.279} & \scalebox{0.78}{0.295} & \scalebox{0.78}{0.273} & \scalebox{0.78}{\textcolor{blue}{0.292}} & \scalebox{0.78}{0.578} & \scalebox{0.78}{0.523} \\
 & \scalebox{0.78}{720} & \scalebox{0.78}{\textcolor{blue}{0.344}} & \scalebox{0.78}{\textcolor{red}{0.343}} & \scalebox{0.78}{0.358} & \scalebox{0.78}{0.349} & \scalebox{0.78}{0.351} & \scalebox{0.78}{\textcolor{blue}{0.344}} & \scalebox{0.78}{0.354} & \scalebox{0.78}{0.348} & \scalebox{0.78}{0.829} & \scalebox{0.78}{0.637} & \scalebox{0.78}{0.926} & \scalebox{0.78}{0.710} & \scalebox{0.78}{0.349} & \scalebox{0.78}{0.345} & \scalebox{0.78}{0.350} & \scalebox{0.78}{0.345} & \scalebox{0.78}{0.398} & \scalebox{0.78}{0.418} & \scalebox{0.78}{\textcolor{red}{0.343}} & \scalebox{0.78}{\textcolor{red}{0.343}} & \scalebox{0.78}{0.351} & \scalebox{0.78}{0.386} & \scalebox{0.78}{0.355} & \scalebox{0.78}{0.347} & \scalebox{0.78}{0.345} & \scalebox{0.78}{\textcolor{blue}{0.344}} & \scalebox{0.78}{1.059} & \scalebox{0.78}{0.741} \\
\cmidrule(lr){2-30}
 & \scalebox{0.78}{Avg} & \scalebox{0.78}{\textcolor{red}{0.243}} & \scalebox{0.78}{\textcolor{red}{0.271}} & \scalebox{0.78}{0.258} & \scalebox{0.78}{0.278} & \scalebox{0.78}{0.255} & \scalebox{0.78}{0.277} & \scalebox{0.78}{0.259} & \scalebox{0.78}{0.280} & \scalebox{0.78}{0.590} & \scalebox{0.78}{0.515} & \scalebox{0.78}{0.657} & \scalebox{0.78}{0.573} & \scalebox{0.78}{0.254} & \scalebox{0.78}{0.279} & \scalebox{0.78}{0.251} & \scalebox{0.78}{0.276} & \scalebox{0.78}{0.259} & \scalebox{0.78}{0.315} & \scalebox{0.78}{\textcolor{blue}{0.246}} & \scalebox{0.78}{0.274} & \scalebox{0.78}{0.270} & \scalebox{0.78}{0.320} & \scalebox{0.78}{0.257} & \scalebox{0.78}{0.276} & \scalebox{0.78}{0.249} & \scalebox{0.78}{\textcolor{blue}{0.273}} & \scalebox{0.78}{0.634} & \scalebox{0.78}{0.548} \\
\midrule
\multirow{5}{*}{\rotatebox{90}{\scalebox{0.95}{Solar-Energy}}} & \scalebox{0.78}{96} & \scalebox{0.78}{\textcolor{red}{0.195}} & \scalebox{0.78}{\textcolor{blue}{0.224}} & \scalebox{0.78}{0.203} & \scalebox{0.78}{0.237} & \scalebox{0.78}{0.230} & \scalebox{0.78}{0.283} & \scalebox{0.78}{0.234} & \scalebox{0.78}{0.286} & \scalebox{0.78}{0.345} & \scalebox{0.78}{0.333} & \scalebox{0.78}{0.386} & \scalebox{0.78}{0.372} & \scalebox{0.78}{0.209} & \scalebox{0.78}{0.260} & \scalebox{0.78}{0.205} & \scalebox{0.78}{0.244} & \scalebox{0.78}{0.310} & \scalebox{0.78}{0.331} & \scalebox{0.78}{0.247} & \scalebox{0.78}{0.292} & \scalebox{0.78}{0.312} & \scalebox{0.78}{0.399} & \scalebox{0.78}{\textcolor{blue}{0.199}} & \scalebox{0.78}{\textcolor{red}{0.221}} & \scalebox{0.78}{--} & \scalebox{0.78}{--} & \scalebox{0.78}{--} & \scalebox{0.78}{--} \\
 & \scalebox{0.78}{192} & \scalebox{0.78}{\textcolor{red}{0.225}} & \scalebox{0.78}{\textcolor{red}{0.252}} & \scalebox{0.78}{0.233} & \scalebox{0.78}{0.261} & \scalebox{0.78}{0.263} & \scalebox{0.78}{0.306} & \scalebox{0.78}{0.267} & \scalebox{0.78}{0.310} & \scalebox{0.78}{0.397} & \scalebox{0.78}{0.366} & \scalebox{0.78}{0.444} & \scalebox{0.78}{0.409} & \scalebox{0.78}{\textcolor{blue}{0.231}} & \scalebox{0.78}{0.269} & \scalebox{0.78}{0.237} & \scalebox{0.78}{0.270} & \scalebox{0.78}{0.734} & \scalebox{0.78}{0.725} & \scalebox{0.78}{0.288} & \scalebox{0.78}{0.320} & \scalebox{0.78}{0.339} & \scalebox{0.78}{0.416} & \scalebox{0.78}{0.237} & \scalebox{0.78}{\textcolor{blue}{0.254}} & \scalebox{0.78}{--} & \scalebox{0.78}{--} & \scalebox{0.78}{--} & \scalebox{0.78}{--} \\
 & \scalebox{0.78}{336} & \scalebox{0.78}{\textcolor{red}{0.245}} & \scalebox{0.78}{\textcolor{red}{0.265}} & \scalebox{0.78}{0.248} & \scalebox{0.78}{\textcolor{blue}{0.273}} & \scalebox{0.78}{0.285} & \scalebox{0.78}{0.310} & \scalebox{0.78}{0.290} & \scalebox{0.78}{0.315} & \scalebox{0.78}{0.427} & \scalebox{0.78}{0.389} & \scalebox{0.78}{0.471} & \scalebox{0.78}{0.429} & \scalebox{0.78}{\textcolor{blue}{0.246}} & \scalebox{0.78}{0.275} & \scalebox{0.78}{0.258} & \scalebox{0.78}{0.288} & \scalebox{0.78}{0.750} & \scalebox{0.78}{0.735} & \scalebox{0.78}{0.329} & \scalebox{0.78}{0.344} & \scalebox{0.78}{0.368} & \scalebox{0.78}{0.430} & \scalebox{0.78}{0.249} & \scalebox{0.78}{\textcolor{blue}{0.273}} & \scalebox{0.78}{--} & \scalebox{0.78}{--} & \scalebox{0.78}{--} & \scalebox{0.78}{--} \\
 & \scalebox{0.78}{720} & \scalebox{0.78}{\textcolor{red}{0.241}} & \scalebox{0.78}{\textcolor{red}{0.268}} & \scalebox{0.78}{\textcolor{blue}{0.249}} & \scalebox{0.78}{0.275} & \scalebox{0.78}{0.287} & \scalebox{0.78}{0.315} & \scalebox{0.78}{0.289} & \scalebox{0.78}{0.317} & \scalebox{0.78}{0.423} & \scalebox{0.78}{0.387} & \scalebox{0.78}{0.473} & \scalebox{0.78}{0.432} & \scalebox{0.78}{0.255} & \scalebox{0.78}{\textcolor{blue}{0.274}} & \scalebox{0.78}{0.260} & \scalebox{0.78}{0.288} & \scalebox{0.78}{0.769} & \scalebox{0.78}{0.765} & \scalebox{0.78}{0.341} & \scalebox{0.78}{0.352} & \scalebox{0.78}{0.370} & \scalebox{0.78}{0.425} & \scalebox{0.78}{0.250} & \scalebox{0.78}{0.276} & \scalebox{0.78}{--} & \scalebox{0.78}{--} & \scalebox{0.78}{--} & \scalebox{0.78}{--} \\
\cmidrule(lr){2-30}
 & \scalebox{0.78}{Avg} & \scalebox{0.78}{\textcolor{red}{0.227}} & \scalebox{0.78}{\textcolor{red}{0.252}} & \scalebox{0.78}{\textcolor{blue}{0.233}} & \scalebox{0.78}{0.262} & \scalebox{0.78}{0.266} & \scalebox{0.78}{0.303} & \scalebox{0.78}{0.270} & \scalebox{0.78}{0.307} & \scalebox{0.78}{0.398} & \scalebox{0.78}{0.369} & \scalebox{0.78}{0.444} & \scalebox{0.78}{0.410} & \scalebox{0.78}{0.235} & \scalebox{0.78}{0.270} & \scalebox{0.78}{0.240} & \scalebox{0.78}{0.273} & \scalebox{0.78}{0.641} & \scalebox{0.78}{0.639} & \scalebox{0.78}{0.301} & \scalebox{0.78}{0.327} & \scalebox{0.78}{0.347} & \scalebox{0.78}{0.417} & \scalebox{0.78}{0.234} & \scalebox{0.78}{\textcolor{blue}{0.256}} & \scalebox{0.78}{--} & \scalebox{0.78}{--} & \scalebox{0.78}{--} & \scalebox{0.78}{--} \\
\midrule
\multirow{5}{*}{\rotatebox{90}{\scalebox{0.95}{PEMS03}}} & \scalebox{0.78}{12} & \scalebox{0.78}{\textcolor{red}{0.065}} & \scalebox{0.78}{\textcolor{red}{0.168}} & \scalebox{0.78}{0.071} & \scalebox{0.78}{0.174} & \scalebox{0.78}{0.095} & \scalebox{0.78}{0.212} & \scalebox{0.78}{0.099} & \scalebox{0.78}{0.216} & \scalebox{0.78}{0.132} & \scalebox{0.78}{0.288} & \scalebox{0.78}{0.139} & \scalebox{0.78}{0.302} & \scalebox{0.78}{--} & \scalebox{0.78}{--} & \scalebox{0.78}{\textcolor{red}{0.065}} & \scalebox{0.78}{\textcolor{blue}{0.169}} & \scalebox{0.78}{0.090} & \scalebox{0.78}{0.203} & \scalebox{0.78}{0.078} & \scalebox{0.78}{0.190} & \scalebox{0.78}{0.178} & \scalebox{0.78}{0.305} & \scalebox{0.78}{\textcolor{blue}{0.067}} & \scalebox{0.78}{\textcolor{blue}{0.169}} & \scalebox{0.78}{--} & \scalebox{0.78}{--} & \scalebox{0.78}{--} & \scalebox{0.78}{--} \\
 & \scalebox{0.78}{24} & \scalebox{0.78}{\textcolor{red}{0.083}} & \scalebox{0.78}{\textcolor{red}{0.190}} & \scalebox{0.78}{0.093} & \scalebox{0.78}{0.201} & \scalebox{0.78}{0.138} & \scalebox{0.78}{0.254} & \scalebox{0.78}{0.142} & \scalebox{0.78}{0.259} & \scalebox{0.78}{0.190} & \scalebox{0.78}{0.346} & \scalebox{0.78}{0.199} & \scalebox{0.78}{0.363} & \scalebox{0.78}{--} & \scalebox{0.78}{--} & \scalebox{0.78}{\textcolor{blue}{0.087}} & \scalebox{0.78}{\textcolor{blue}{0.196}} & \scalebox{0.78}{0.121} & \scalebox{0.78}{0.240} & \scalebox{0.78}{0.119} & \scalebox{0.78}{0.236} & \scalebox{0.78}{0.257} & \scalebox{0.78}{0.371} & \scalebox{0.78}{0.102} & \scalebox{0.78}{0.213} & \scalebox{0.78}{--} & \scalebox{0.78}{--} & \scalebox{0.78}{--} & \scalebox{0.78}{--} \\
 & \scalebox{0.78}{48} & \scalebox{0.78}{\textcolor{red}{0.116}} & \scalebox{0.78}{\textcolor{red}{0.228}} & \scalebox{0.78}{\textcolor{blue}{0.125}} & \scalebox{0.78}{\textcolor{blue}{0.236}} & \scalebox{0.78}{0.205} & \scalebox{0.78}{0.315} & \scalebox{0.78}{0.211} & \scalebox{0.78}{0.319} & \scalebox{0.78}{0.285} & \scalebox{0.78}{0.431} & \scalebox{0.78}{0.295} & \scalebox{0.78}{0.447} & \scalebox{0.78}{--} & \scalebox{0.78}{--} & \scalebox{0.78}{0.133} & \scalebox{0.78}{0.243} & \scalebox{0.78}{0.202} & \scalebox{0.78}{0.317} & \scalebox{0.78}{0.216} & \scalebox{0.78}{0.329} & \scalebox{0.78}{0.379} & \scalebox{0.78}{0.463} & \scalebox{0.78}{0.169} & \scalebox{0.78}{0.281} & \scalebox{0.78}{--} & \scalebox{0.78}{--} & \scalebox{0.78}{--} & \scalebox{0.78}{--} \\
 & \scalebox{0.78}{96} & \scalebox{0.78}{\textcolor{red}{0.152}} & \scalebox{0.78}{\textcolor{red}{0.262}} & \scalebox{0.78}{\textcolor{blue}{0.164}} & \scalebox{0.78}{\textcolor{blue}{0.275}} & \scalebox{0.78}{0.266} & \scalebox{0.78}{0.365} & \scalebox{0.78}{0.269} & \scalebox{0.78}{0.370} & \scalebox{0.78}{0.360} & \scalebox{0.78}{0.495} & \scalebox{0.78}{0.377} & \scalebox{0.78}{0.518} & \scalebox{0.78}{--} & \scalebox{0.78}{--} & \scalebox{0.78}{0.201} & \scalebox{0.78}{0.305} & \scalebox{0.78}{0.262} & \scalebox{0.78}{0.367} & \scalebox{0.78}{0.357} & \scalebox{0.78}{0.431} & \scalebox{0.78}{0.490} & \scalebox{0.78}{0.539} & \scalebox{0.78}{0.417} & \scalebox{0.78}{0.472} & \scalebox{0.78}{--} & \scalebox{0.78}{--} & \scalebox{0.78}{--} & \scalebox{0.78}{--} \\
\cmidrule(lr){2-30}
 & \scalebox{0.78}{Avg} & \scalebox{0.78}{\textcolor{red}{0.104}} & \scalebox{0.78}{\textcolor{red}{0.212}} & \scalebox{0.78}{\textcolor{blue}{0.113}} & \scalebox{0.78}{\textcolor{blue}{0.222}} & \scalebox{0.78}{0.176} & \scalebox{0.78}{0.286} & \scalebox{0.78}{0.180} & \scalebox{0.78}{0.291} & \scalebox{0.78}{0.242} & \scalebox{0.78}{0.390} & \scalebox{0.78}{0.253} & \scalebox{0.78}{0.408} & \scalebox{0.78}{--} & \scalebox{0.78}{--} & \scalebox{0.78}{0.122} & \scalebox{0.78}{0.228} & \scalebox{0.78}{0.169} & \scalebox{0.78}{0.282} & \scalebox{0.78}{0.193} & \scalebox{0.78}{0.296} & \scalebox{0.78}{0.326} & \scalebox{0.78}{0.419} & \scalebox{0.78}{0.189} & \scalebox{0.78}{0.284} & \scalebox{0.78}{--} & \scalebox{0.78}{--} & \scalebox{0.78}{--} & \scalebox{0.78}{--} \\
\bottomrule
\end{tabular}
\end{small}
\end{threeparttable}
}
\vspace{-0.2in}
\label{tab_all_res}
\end{table*}

\begin{figure}[htbp]
    \centering
    \includegraphics[width=0.80\linewidth]{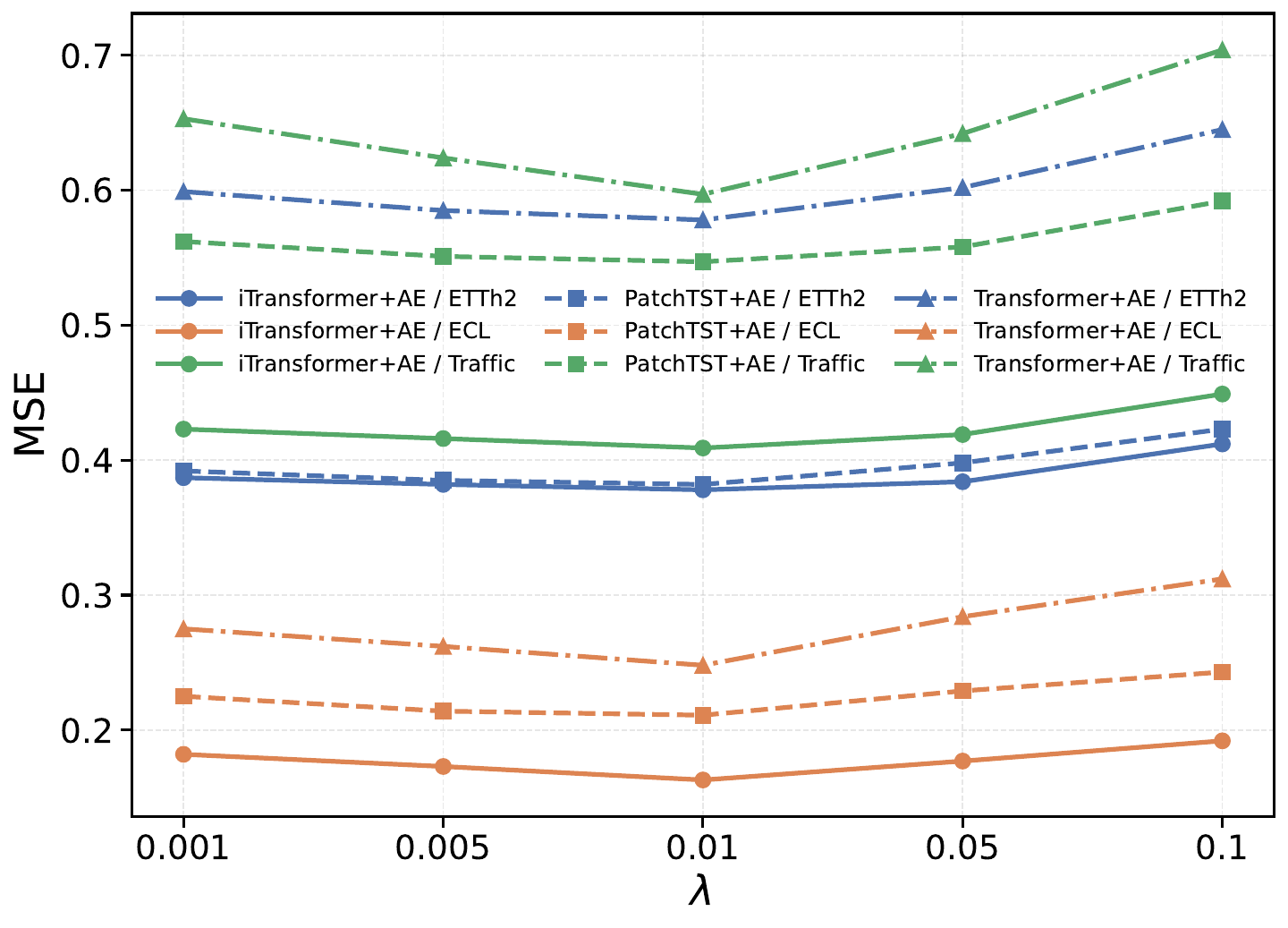}
    \caption{Hyperparameter sensitivity analysis of $\lambda$.}
    \label{fig_sense}
\end{figure} 

\begin{table}[t]
\centering
\caption{Comparison of different attention constraints.}
\label{tab:attention_constraint}
\setlength{\tabcolsep}{4pt}
\renewcommand{\arraystretch}{0.95}

\resizebox{0.7\linewidth}{!}{
\begin{tabular}{lcc}
\toprule
Method & ECL & Traffic \\
\midrule
iTransformer & 0.178 & 0.428 \\
iTransformer + AE & \textbf{0.163} & \textbf{0.410} \\
iTransformer + Fixed Temp. & 0.174 & 0.425 \\
iTransformer + Top-K Attn. & 0.185 & 0.434 \\
iTransformer + $L_1$ Mask & 0.170 & 0.420 \\
\bottomrule
\end{tabular}
}

\end{table}

\section{Experiments}
\subsection{Experimental Settings}
\textbf{Datasets:} We use six real-world datasets: ECL, ETTh2, Traffic, Solar-Energy, PEMS03, and Weather from \cite{liu2024itransformer}. All datasets follow the splits in \cite{liu2024itransformer}.

\textbf{Baselines:} We compare with ten classic TSF methods, including transformer-based approaches (PatchTST \cite{PatchTST}, iTransformer \cite{liu2024itransformer}, Crossformer \cite{Crossformer}, DSformer \cite{yu2023dsformer}, DPAD \cite{yang2026dual}, ARCausal \cite{zhou2026causal}, Informer \cite{zhou2021informer}),  other types of methods (CycleNet/Linear\cite{lin2024cyclenet}, S-Mamba \cite{wang2025mamba}, and TIDE \cite{das2023long}). 

\textbf{Implementation Details:} 
We apply the proposed method to three representative Transformer-based TSF methods (iTransformer \cite{liu2024itransformer}, Transformer \cite{Transformer}, and PatchTST \cite{PatchTST}). The reasons for choosing these three methods are twofold: first, all methods have simple structures that easily integrate with our method; second, they are highly versatile, and many other methods \cite{Crossformer, jiang2023pdformer, yu2023dsformer} are built upon them. The successful application of our method on the three methods suggests that it can be utilized in more models based on them. The proposed models are trained using NVIDIA 4090 GPUs. MSE and MAE are employed as evaluation metrics. The value of $\lambda$ in Eq.\ref{eq_object} is set to 0.01. All other hyperparameters and training settings follow the suggested configurations of the corresponding methods of the official code of \cite{liu2024itransformer}.

\subsection{Comparative Experimental Results}
\label{sec_res}
Table \ref{tab_all_res} presents the forecasting results of all methods, where the best and second-best results are highlighted in red and blue, respectively. Lower MSE and MAE values indicate better forecasting performance. As shown in the table, although the backbone models used in this work are not the current state-of-the-art methods, incorporating AE enables them to match or even outperform more sophisticated methods on multiple datasets, further demonstrating the effectiveness of the proposed AE method.

\subsection{Visualization Experiment}
In this section, we repeat the experiment in Fig.\ref{fig_motivate} after applying our Attention Entropy Constraint (AE). The results are shown in Fig.\ref{visualize}. Comparing Figs.\ref{fig_motivate} and \ref{visualize}, we observe that our method enables the base models to rely on fewer redundant dependencies (i.e., smaller purple regions) while achieving lower prediction MSE. These results demonstrate the effectiveness of our method.

\subsection{Hyperparameter Sensitivity and Ablation}
We first determine the value of $\lambda$ through a hyperparameter sensitivity study. As shown in Fig.~\ref{fig_sense}, the best performance is achieved at $\lambda=0.01$. In addition, we conduct ablation studies on different attention sparsification strategies, with the results reported in Table~\ref{tab:attention_constraint}. Specifically, +Top-K Attn retains only the top $K\%$ token dependencies with the highest attention weights, where $K$ is selected empirically; +$L_1$ Mask applies a sparse mask to the attention matrix, whose sparsity is encouraged by $L_1$ regularization.

\section{Conclusion}
This paper reveals that Transformer-based time series forecasting (TSF) models may exploit redundant token dependencies during forecasting, which can negatively affect prediction performance. To address this issue, we propose a simple yet effective token dependency selection strategy. By jointly optimizing the attention entropy constraint and forecasting loss, we encourage the model to make accurate predictions based on fewer but more critical token dependencies, thereby reducing the interference of redundant dependencies. Extensive experiments validate the effectiveness of the proposed method.

\clearpage
\bibliographystyle{IEEEbib}
\bibliography{strings,refs}


\end{document}